%% file: main.tex
\documentclass[runningheads]{llncs}
\usepackage[T1]{fontenc}
\input{sections/__preamble}

\usepackage{graphicx}
\usepackage{color}

\begin{document}
\title{The Anatomy of Adversarial Attacks:\\ Concept-based XAI Dissection}
%

\author{
Georgii Mikriukov\inst{1,2}\orcidID{0000-0002-2494-6285}
\and Gesina Schwalbe\inst{3}\orcidID{0000-0003-2690-2478}
\and Franz Motzkus\inst{1}\orcidID{0009-0009-4362-7907}
\and Korinna Bade\inst{2}\orcidID{0000-0001-9139-8947}
}

\authorrunning{G. Mikriukov et al.}

\institute{
Continental AG, Germany\\
\email{\{firstname.lastname\}@continental-corporation.com}\\
\and
Hochschule Anhalt, Germany\\
\email{\{firstname.lastname\}@hs-anhalt.de}
\and
University of Lübeck, Germany\\
\email{\{firstname.lastname\}@uni-luebeck.de}
}

%
\maketitle              
%
\input{sections/0_abstract.tex}

\input{sections/1_intro.tex}

\input{sections/2_related.tex}

\input{sections/3_background.tex}

\input{sections/4_setup.tex}

\input{sections/5_experiments.tex}

\input{sections/6_conclusion.tex}

\input{sections/7_acknowledgments.tex}

\bibliographystyle{splncs04}
\bibliography{ref}


\end{document}

%% file: sections/__preamble.tex
\usepackage[hidelinks]{hyperref}
\usepackage{booktabs}
\usepackage{algorithm}
\usepackage{algorithmic}
\usepackage{mathtools}
\usepackage{enumitem}
\usepackage{amssymb,dsfont}
\usepackage{tabularx}
\usepackage{multirow}

\usepackage{caption}
\usepackage{subcaption}
\usepackage{comment}
\usepackage{graphicx}
\usepackage{csquotes}
\usepackage{enumitem}
\usepackage{nth}
\usepackage{makecell}
\usepackage{bm} 

\usepackage{cleveref}

\usepackage{xcolor}
\definecolor{plotBlack}{RGB}{0,0,0}
\definecolor{plotRed}{RGB}{255,0,0}
\definecolor{plotBlue}{RGB}{0,0,255}
\definecolor{plotGreen}{RGB}{0,128,0}

\newcommand{\blackLine}{\texttt{\textcolor{plotBlack}{black}}}
\newcommand{\redLine}{\texttt{\textcolor{plotRed}{red}}}
\newcommand{\blueLine}{\texttt{\textcolor{plotBlue}{blue}}}
\newcommand{\greenLine}{\texttt{\textcolor{plotGreen}{green}}}

\renewcommand{\paragraph}[1]{\par\noindent\textbf{#1}\,}
\newcommand*{\Reals}{\mathds{R}}

\makeatletter
\usepackage{tikz}
\newcommand*\circled[2][2pt]{\tikz[baseline=(char.base)]{
    \node[shape=circle, draw, inner sep=#1, 
        ]
        (char) {#2};}}
\makeatother

\newcommand{\ResearchQuestion}[1]{%
\paragraph{\circled{\sffamily\textbf{?}} Research Question:}%
\textit{#1}%
}

\newcommand{\Summary}[1]{%
\paragraph{\circled[.5pt]{$\bigstar$} Summary:}%
\textit{#1}%
}

%% file: sections/0_abstract.tex
\begin{abstract}

Adversarial attacks (AAs) pose a significant threat to the reliability and robustness of deep neural networks.
While the impact of these attacks on model predictions has been extensively studied, their effect on the learned representations and concepts within these models remains largely unexplored. In this work, we perform an in-depth analysis of the influence of AAs on the concepts learned by convolutional neural networks (CNNs) using eXplainable artificial intelligence (XAI) techniques.
Through an extensive set of experiments across various network architectures and targeted AA techniques, we unveil several key findings.
First, AAs induce substantial alterations in the concept composition within the feature space, introducing new concepts or modifying existing ones.
Second, the adversarial perturbation itself can be linearly decomposed into a set of latent vector components, with a subset of these being responsible for the attack's success. 
Notably, we discover that these components are target-specific, i.e., are similar for a given target class throughout different AA techniques and starting classes.
%
Our findings provide valuable insights into the nature of AAs and their impact on learned representations, paving the way for the development of more robust and interpretable deep learning models, as well as effective defenses against adversarial threats.

\keywords{Concept-based XAI \and Concept Activation Vectors \and Concept Discovery \and Adversarial Attack \and Security}

\end{abstract}


%% file: sections/1_intro.tex
\section{Introduction}
\label{sec:intro}

Deep neural networks (DNN), in particular convolutional DNNs (CNNs), have achieved remarkable success in 
computer vision~\cite{zhao2019object},
but are known to be vulnerable to adversarial evasion attacks (AA) \cite{akhtar2021advances,szegedy2013intriguing,ilyas2019adversarial,madry2017towards}: For a given input an attacker can easily find targeted, small perturbations of the input that are seemingly irrelevant \cite{brown2017adversarial} or even imperceptible \cite{ilyas2019adversarial} to humans, but strongly change the CNN output \cite{goodfellow2014explaining,szegedy2013intriguing}, possibly causing erroneous outputs. CNN vulnerability to AAs has raised significant concerns regarding their reliability and robustness~\cite{szegedy2013intriguing,goodfellow2014explaining}. 
%
While the impact of AAs on model predictions has been extensively studied~\cite{%
akhtar2021advances,
madry2017towards,
goodfellow2014explaining,
kurakin2018adversarial
}, their effect on the internal representations 
learned by these models remains largely unexplored. Understanding how AAs influence the learned representations is crucial for developing robust CNNs, as well as effective defenses against adversarial threats.

Recent advancements in XAI 
have provided valuable tools for probing the internal representations learned by deep neural networks~\cite{%
schwalbe_comprehensive_2023,
schwalbe2022concept,
kim2018interpretability,
fong2018net2vec
}. In particular, XAI methods for unsupervised concept embedding analysis \cite{schwalbe2020concept} have proven successful in identifying frequent patterns in CNN latent representations that are used for the CNN's decision-making~\cite{%
zhang2021invertible,
ghorbani2019towards,
ge2021peek
}. These reoccurring patterns respectively linear components are referred to as concepts.

In this work, we perform an in-depth analysis of the influence of strong targeted AAs on the usage of concepts learned by CNNs using concept-based XAI techniques. Inspired by the perspective of Ilyas et.\ al~\cite{ilyas2019adversarial} that \enquote{adversarial examples are no bugs, they are features}, we hypothesize that AAs exploit and manipulate the learned concepts within latent spaces of these models. Through a comprehensive set of experiments across various network architectures and targeted strong white-box AA types, we unveil several key findings that shed light on the nature of AAs and their impact on learned representations:

\begin{enumerate}[label=(\arabic*)]
    \item \textbf{Adversarial attacks substantially alter the concept composition} learned by CNNs, introducing new concepts or modifying existing ones.
    \item \textbf{Adversarial perturbations can be decomposed into linear components}, only a subset of which is primarily responsible for the attack's success.
    \item Different attacks comprise similar components, suggesting they nudge the CNN intermediate outputs towards specific \textbf{common adversarial feature space directions}, albeit with varying magnitudes.
    \item The learned adversarial concepts are mostly \textbf{specific to the target class}, agnostic of the class the attack starts from. This implies that attacks abuse target-specific feature space directions.
\end{enumerate}
These insights into AAs' impact on learned representations open up new directions for the future design of more robust models and adversarial defenses.


The remainder of the paper is structured as follows:
After setting this paper in the context of related work (Sec.\,\ref{sec:related}),
we introduce the necessary background on used AA and XAI techniques (Sec.\,\ref{sec:background}),
and the common experimental setup (Sec.\,\ref{sec:setup}) for investigating our hypothesis.
Section\,\ref{sec:experiments} then presents the experimental studies and their results.
We conclude in Section\,\ref{sec:conclusion} with a discussion of implications for the understanding and handling of AAs.


%% file: sections/2_related.tex
\section{Related Work}
\label{sec:related}


\paragraph{Adversarial attacks}
\label{sec:related-attacks}
%
AAs on CNNs have received increased attention \cite{akhtar2021advances} since the first description of the phenomenon in 2014 \cite{%
szegedy2013intriguing,
goodfellow2014explaining
}. These attacks generate perturbations to input data that can lead to erroneous model predictions with high confidence~\cite{%
goodfellow2014explaining
}. 
%
Numerous attack methods have been proposed, ranging 
from digital to physically implemented manipulations \cite{eykholt2018robust}, 
and from white-box attacks against fully known models to purely query-based ones \cite{akhtar2021advances}. Commonly used strong AAs, i.e., ones achieving the largest output deviation at minor input changes, are the here considered white-box attacks \cite{akhtar2021advances,carlini2017towards}. First of its kind was the gradient-based Fast Gradient Sign Method (FGSM)~\cite{goodfellow2014explaining}, refined to the Basic Iterative Method (BIM)~\cite{kurakin2018adversarial}, and Projected Gradient Descent (PGD)~\cite{madry2017towards}. These were later superseded by the optimization-based Carlini and Wagner (C\&W)~\cite{carlini2017towards} attack.
Another common type of attack is adversarial patch~\cite{brown2017adversarial}, which aims to simulate real-world scenarios by restraining perturbations to be local. This allows them to be physically realized, e.g., via stickers \cite{eykholt2018robust}.

Considerable efforts were invested to develop defenses against adversarial attacks \cite{bai_recent_2021,akhtar2021advances}.
However, many of these are still vulnerable to adaptive attacks~\cite{athalye2018obfuscated,carlini_adversarial_2017}, highlighting the ongoing arms race between attacks and defenses.
Meanwhile, the underlying nature and causes of adversarial vulnerabilities is not satisfyingly clarified: Hypotheses include that they origin from invisible, class-correlated features in the data \cite{ilyas2019adversarial}, high-frequency image features \cite{varghese_unsupervised_2021}, high dimensionality \cite{ghorbani2019interpretation} and curvature \cite{dombrowski2019explanations} of CNN latent spaces, and error amplification along singular vulnerable latent features \cite{madaan_adversarial_2020}. But none could be proven to cover all cases of adversaries, leaving the nature of AAs an open question.

\paragraph{Concept-based XAI Methods}
\label{sec:related-concepts}
While it might be more desirable to use fully transparent models \cite{rudin_stop_2019}, black-box CNNs are hard to replace for vision tasks \cite{zhao2019object}.
Fortunately, post-hoc XAI methods allow to explain many aspects of a trained black-box CNN \cite{schwalbe_comprehensive_2023}, including, e.g., the \emph{importance of input features} for decisions \cite{%
samek2019explainable
}, more general \emph{model behavior} (using simplified surrogates) \cite{rabold_explaining_2018,chyung_extracting_2019}, and internal \emph{information flow} \cite{hohman_summit_2020}.
The question \emph{what information is encoded in CNN intermediate outputs} is tackled by the subfield of concept-based XAI methods: Here, CNN latent space structures are associated with human-understandable symbolic concepts \cite{schwalbe_concept_2022}.
%
Early works started to explain the meaning of single CNN units \cite{bau2017network,olah2017feature}. Kim et al.\ \cite{kim2018interpretability} soon after found that more generally vectors respectively directions in CNN latent space are a favorable target for explaining the information encoded in CNN intermediate outputs
(other than non-linear approaches \cite{esser_disentangling_2020}).
Linear approaches nowadays either work \emph{supervised}, i.e., find concept vectors for user-pre-defined concepts \cite{kim2018interpretability,fong2018net2vec}; or, as considered here, \emph{unsupervised}, by extracting main linear components from CNN intermediate outputs, to then visualize and interactively interpret them \cite{zhang2021invertible,ghorbani2019towards}.
Note that, unlike supervised concept-based analysis, these patterns, i.e., concepts, are not always human-interpretable. The concept discovery is commonly achieved through techniques such as clustering of activations~\cite{ghorbani2019towards}, activation pixels~\cite{posada2022eclad}, and---more effective---matrix decomposition methods like NMF~\cite{zhang2021invertible,fel2023craft}, PCA~\cite{zhang2021invertible,fel2023craft}, ICA~\cite{fel2023craft} or RCA~\cite{fel2023craft}. This study will focus on NMF and PCA, as the former provides more interpretable concepts, while the latter offers greater numerical efficiency~\cite{zhang2021invertible}.
Visualization of a component is done by highlighting spatial input regions where the component is prominent in CNN activation map locations. That way, unsupervised concept-based approaches allow us to gain insights into the semantic features the CNN has learned to use,
enabling us to observe changes in used knowledge at the semantic level. This is here leveraged to assess the influence of adversarial attacks on CNN knowledge.

\paragraph{Adversarial Attacks and XAI}
\label{sec:related-aa-vs-xai}
%
%
%
While adversarial attacks and XAI techniques have been extensively studied, their intersection has received limited attention. Recent interdisciplinary works have only explored the use of feature importance XAI methods
for the detection of adversarial samples~\cite{garcia2018explainable,fidel2020explainability,kao2022rectifying,serrurier2022adversarial,chen2022adversarial,rieger2020simple}; and showed, that both feature importance \cite{dombrowski2019explanations,ghorbani2019interpretation} and concept representations \cite{brown2023making} can be attacked, i.e., modified in a targeted manner via input perturbations.
%
%
However, when targeting CNN outputs, the impact of AAs on the learned representations and concepts within deep neural networks remains largely unexplored. First investigations on the nature of AAs only investigated origins in input features \cite{ilyas2019adversarial}. Later ones, going inside the CNN, were restricted to overall structural properties like dimensionality \cite{ghorbani2019interpretation} and curvature \cite{dombrowski2019explanations}, or the impact of singular neurons to amplify errors \cite{madaan_adversarial_2020}.
To our knowledge, we are the first to take a look at the interplay between AAs and learned concepts in CNN latent spaces.



%% file: sections/3_background.tex
\section{Background}
\label{sec:background}

In this section, we provide details of adversarial attacks (Sec.~\ref{sec:background-attacks}), concept discovery (Sec.~\ref{sec:background-concept-discovery}), and concept comparison methods (Sec.~\ref{sec:background-comparison}) used in this work.

\subsection{Adversarial Attacks}
\label{sec:background-attacks}

Given an input image $x\in\Reals^{h\times w}$ and a classification model $f\colon \Reals^{h\times w}\to Y$ with original prediction $y = f(x)\in Y$, the primary objective of an adversarial attack is to discover or generate a perturbation $\delta$ of maximum size $\|\delta\|\leq \epsilon$ such that the output of $f$ on the adversarial example $x+\delta$ changes ($f(x+\delta)\neq y$, respectively the output difference $\|f(x+\delta)-f(x)\|$ is large for chosen metric $\|\cdot\|$).
The size constraint ensures that $x$ and its perturbed version $x+\delta$ are similar respectively indifferentiable for humans.


In this work, we aim to investigate the impact of adversarial attacks on the level of concepts, and in particular relative to the AA's impact on the DNN output. As this requires fine-grained control on the severity and direction of attacks, we here focus on \emph{targeted white-box optimization-based attacks}.
\emph{Targeted} attacks aim to manipulate the model's prediction from the true class $y$ to a specific pre-defined target class $y'$, thus ensuring $f(x + \delta) = y'$.
\emph{White-box} attacks utilize model internal information like gradients to generate the perturbation, and \emph{optimization-based} ones optimize from $x$ to $x+\delta$ in a controllable step-wise manner.
Our utilized attacks cover a wide range of common white-box attack techniques explained in the following: BIM~\cite{kurakin2018adversarial}, PGD~\cite{madry2017towards}, C\&W~\cite{carlini2017towards} and adversarial patch \cite{brown2017adversarial}, visualized in Fig.~\ref{fig:adversarial-sample-examples}.

\input{graphics/figure_adversarial_samples_example}

\subsubsection{BIM}
The Basic Iterative Method (BIM)~\cite{kurakin2018adversarial} is an iterative variant of the original FGSM attack~\cite{goodfellow2014explaining}. In BIM, adversarial perturbations are iteratively computed by changing $x$ in small steps in direction of the gradient until $x+\delta$ is misclassified. Formally:
\begin{align}
\label{eq:attack-bim}
    x^{t+1} &= \text{clip}_{\epsilon}\left(x^{t} + \alpha \cdot \text{sign}(\nabla_x J(f(x^{t}), y'))\right)
\end{align}
where $t$ represents the optimization step, $J(\cdot)$ denotes the cost function (e.g., cross-entropy), $\nabla_x$ the gradient at $x$, $\alpha$ is a step size hyperparameter, $\text{sign}(\cdot)$ is the element-wise signum function, and $\text{clip}_{\epsilon}(\cdot)$ clips $x+\delta$ element-wise such that it lies in the $L_\infty$ ball of radius $\epsilon$ around $x$.

\subsubsection{PGD}
Another powerful iterative attack is the Projected Gradient Descent (PGD)~\cite{madry2017towards}. Unlike the value clipping approach used in BIM, the PGD attack projects gradients to the $L^{\infty}$ $\epsilon$-ball around the original image:
\begin{align}
    \label{eq:attack-pgd}
    x^{t+1} &= \text{proj}_{\epsilon}\left(x^{t} + \alpha \cdot \text{sign}(\nabla_x J(f(x^{t}), y'))\right)
\end{align}
where $\text{proj}_{\epsilon}$ denotes the projection onto the $\epsilon$-ball centered at $x$.

\subsubsection{C\&W}
The Carlini \& Wagner (C\&W)~\cite{carlini2017towards} attack aims to find the smallest perturbation $\delta$ that leads to misclassification. It formulates the attack as an optimization problem, using the $L_p$ norm (typically $L_2$) to measure the size of the perturbation. Formally, given a helper function $h(\cdot)$ fulfilling $h(x')\leq 0 \Leftrightarrow f(x')=y'$, they optimize:
\begin{gather}
  \SwapAboveDisplaySkip
  \min_{\delta} \| \delta \|_p + \beta \cdot h(x + \delta)
  \label{eq:attack-cw}
\end{gather}
where $\| \cdot \|_p$ denotes the $L_p$ norm, and $\beta$ is a hyperparameter that controls the trade-off between the size of the perturbation and the induced output change. $h(\cdot)$ can be set to, e.g., $h(x')=1-J(f(x'),y')$, where $J(\cdot)$ is the cross-entropy.

\subsubsection{Adversarial Patch}

An adversarial patch \cite{brown2017adversarial} is the most common real-world attack \cite{akhtar2021advances,eykholt2018robust,brown2017adversarial}. Here, $\delta$ is constrained to be local to a region $\textit{loc}$, i.e., only change pixels within $\textit{loc}$.
The induced input changes may be visible to humans, but are easily overlooked as being irrelevant, and can be physically implemented (e.g., as stickers \cite{eykholt2018robust}), cf.~Fig.\,\ref{fig:adversarial-sample-examples}.
The choice of the patch $\delta^{loc}$ and its location $loc$ on the image can vary depending on the specific objectives of the attack~\cite{akhtar2021advances}.
%

\subsection{Concept Discovery with Matrix Factorization}
\label{sec:background-concept-discovery}


For the concept discovery with matrix factorization, the main assumption is that CNNs learn few relevant concepts which can be represented as a linear combination of convolutional filters~\cite{fong2018net2vec,zhang2021invertible}. I.e., \emph{concept-related information is distributed across channels} of sample activation maps.
For the formalization assume that we are given a set $X$ of $b$ probing input images with activations $A=f_{\to L}(X) \in \Reals^{(b \times h \times w) \times c}$ in a selected layer $L$ of the tested CNN $f$, where $b$, $h$, $w$, and $c$ are batch, height, width, and channel dimensions respectively.
The decomposition techniques aim to find a matrix $M  \in \Reals^{k \times c}$ of $k$ components/concepts that allows to (approximately) write each activation map pixel $a\in\Reals^c$ in $A$ as a linear combination of the components in $M$.
Formally, one optimizes the concepts $M$ and weights $W\in \Reals^{(b \times h \times w) \times k}$ such that the reconstruction error is minimized with respect to the Frobenius norm:
%
\begin{align}
    \min_{W, M}\|A - W M\|
    \label{eq:decomposition}
\end{align}
%
To visualize how concepts are activated in new images, one can project concept information back to the input: Assume we already have concept components $M$ from optimization on a concept probing set $A$. Now we want to find for a new activation map ${A}'\in\Reals^{(1\times h\times w)\times c}$ which image region activated which concepts. To do so we
follow \cite{zhang2021invertible,fel2023craft,mikriukov2023quantified} and
obtain $W'\coloneqq A' \cdot M$.
Resulting ${W}'\in\Reals^{(1\times h\times w)\times k}$ holds for each of the $k$ components a \emph{concept saliency map} of size $h\times w$ that tells at which activation map pixel how much of the component was part of the pixel vector. Using the spatial alignment of activation map pixels with the input, one can scale such concept saliency maps from any layer to match the input and, e.g., visualize them as overlays (cf.~Fig.\,\ref{fig:concept-mining-555-468}).

\subsubsection{PCA} Here, one uses mean-centered activations $A_\odot = A - \mu_{A}$, where $\mu_{A}$ is the mean of $A$, ensuring that the found principal components represent the directions of maximum variance in the data. In PCA, the component tensor $M^{\text{PCA}}$ represents the top-$k$ largest eigenvectors of the covariance matrix of $A_\odot$. These eigenvectors capture the directions of maximum variance in the data. The factorization reads: 
\begin{align}
\label{eq:decomposition-pca}
    A_\odot &\approx W M^{\text{PCA}}
%
\end{align}

\subsubsection{NMF} A fundamental constraint of NMF is the non-negativity of activations, i.e., it works on $A^+$ (with $(\cdot)^+\coloneqq \texttt{ReLU}(\cdot)$, where $\texttt{ReLu}(z) = \max(0, z)$), and yields components and weights with non-negative entries ($M=M^+, W=W^+$). This constraint aligns with the interpretation that the $k$ components and should represent additive contributions. Formally:
\begin{align}
    \SwapAboveDisplaySkip
    A^+ \approx W^+ M^+
    \label{eq:decomposition-nmf}
\end{align}

\subsection{Concept Comparison}
\label{sec:background-comparison}

Comparison of concepts is important for the estimation of the similarity of knowledge they represent.
Concepts within the same feature space can be effectively compared using various vector operations:
cosine similarity~\cite{kim2018interpretability,mikriukov2023evaluating}, 
vector arithmetic~\cite{fong2018net2vec}, 
and distance-based metrics~\cite{crabbe2022concept,zhang2021invertible,posada2022eclad,mikriukov2023gcpv}. 
%
In this work, due to the nature of concepts --- a linear combination of convolutional filters --- obtained from decomposition-based concept discovery methods, we utilize cosine similarity for comparing any two discovered concepts $u, v$:
\begin{align}
\text{sim} ( u, v ) = \frac{u\cdot v}{\|u\|\cdot \|v\|}
\label{eq:cosine-similarity}
\end{align}
This ensures that only the direction of vectors is considered for the comparison.

When comparing concepts obtained from different layers and/or models, i.e., in distinct latent spaces, one needs to project the concept information to a common space, such as input or output. One output-based approach are metrics for estimating the attribution of concepts to the CNN outputs. These metrics include gradient-based approaches \cite{kim2018interpretability,crabbe2022concept}, 
saliency-based methods \cite{mikriukov2023quantified}, 
or those based on concept similarity ranking \cite{mikriukov2023quantified}. 
We here compare concepts quantitatively using the Jaccard Index (IoU) of their concept saliency maps after scaling them to match the input size, as proposed in~\cite{mikriukov2023quantified}.





%% file: graphics/figure_adversarial_samples_example.tex
\begin{figure*}[ht]
  \centering
  \includegraphics[width=1.0\linewidth]{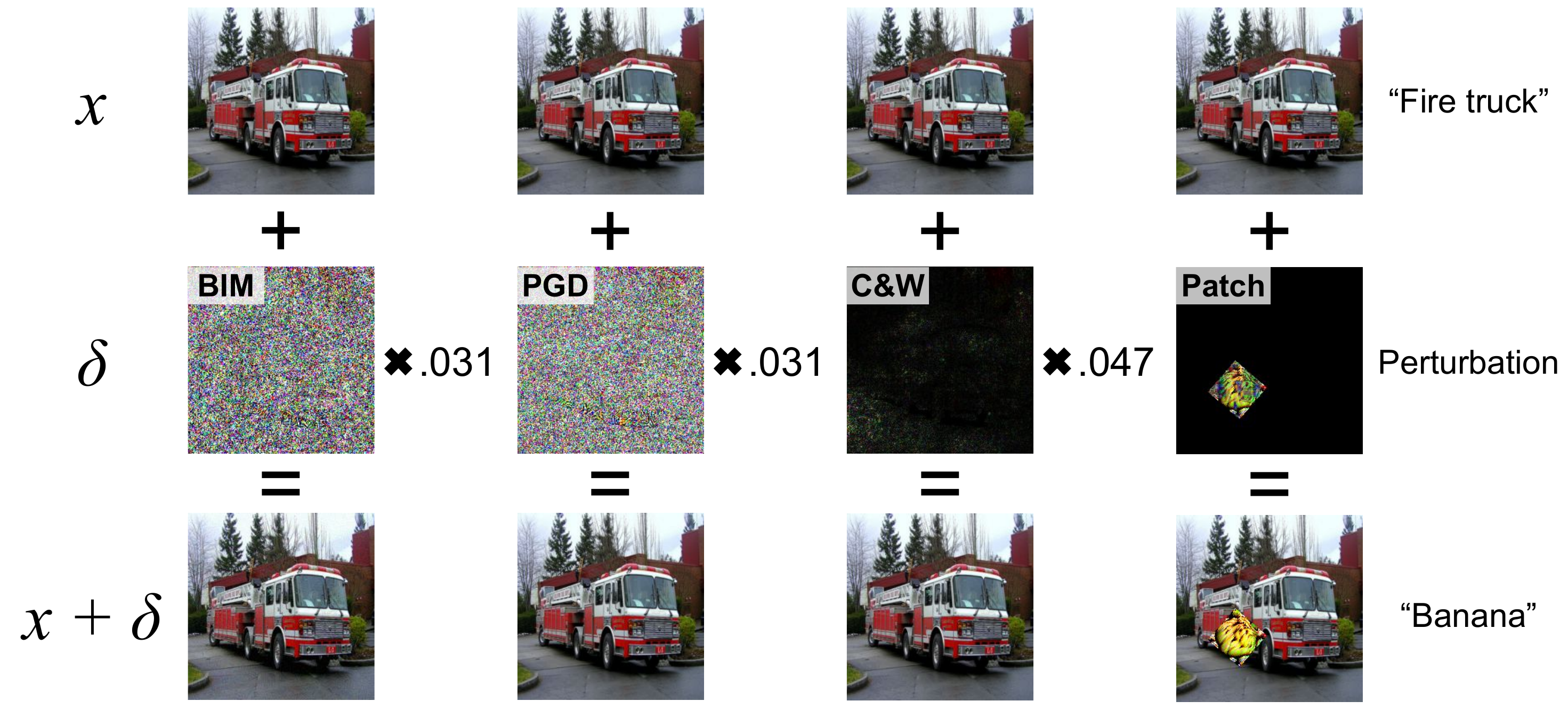}
  \caption{Examples of BIM, PGD, C\&W, and Patch Attack adversarial samples: \enquote{fire truck} attacked with target \enquote{banana}.}
  \label{fig:adversarial-sample-examples}
\end{figure*}

%% file: sections/4_setup.tex
\section{Experimental Setup}
\label{sec:setup}

In this section we present the selection of models (Sec.~\ref{sec:setup-models}), data used in experiments (Sec.~\ref{sec:setup-data}), and layers selection for the analysis of concepts and internal representations (Sec.~\ref{sec:setup-layers}).

\subsection{Models}
\label{sec:setup-models}
In our experiments we use classification models of different architectures trained on ImageNet~\cite{deng2009imagenet} dataset from PyTorch 
model zoo\footnote[1]{\url{https://pytorch.org/vision/stable/models\#classification}}: 
\begin{itemize}
    \item VGG-11~\cite{simonyan2014very} (\emph{VGG})
    \item Compressed SqueezeNet1.1~\cite{iandola2016squeezenet} (\emph{SqueezeNet}),
    \item Inverted residual MobileNetV3-Large~\cite{howard2019searching} (\emph{MobileNet})
    \item Residual ResNet18~and~ResNet50~\cite{he2016deep}. Two networks of the same architecture are used to investigate the impact of the model size.
\end{itemize}
All models are pre-trained on the ImageNet1k~\cite{deng2009imagenet} dataset.

\subsection{Data}
\label{sec:setup-data}
We conduct our experiments using diverse classes from a validation subset of ILSVRC2017~\cite{deng2009imagenet}. Specifically, we selected four classes from the \texttt{vehicle} supercategory (\texttt{taxi}, \texttt{fire truck}, \texttt{garbage truck}, \texttt{pickup truck}), two classes from the \texttt{animal} supercategory (\texttt{horse}, \texttt{zebra}), and two classes from the \texttt{fruit} supercategory (\texttt{orange}, \texttt{banana}).

For each of the selected classes, we chose 50 images, and for each network subjected them to targeted cross-attacks using gradient-based BIM~\cite{kurakin2018adversarial}, PGD~\cite{madry2017towards}, and C\&W~\cite{carlini2017towards} (see Sec.~\ref{sec:background-attacks}). Gradient attacks were executed using \texttt{torchattacks}~\cite{kim2020torchattacks} library\footnote[2]{\url{https://github.com/Harry24k/adversarial-attacks-pytorch}}. Additionally, we used the ImageNet-Patch\footnote[3]{\url{https://github.com/pralab/ImageNet-Patch}}~\cite{pintor2023imagenet} dataset, which comprises pretrained adversarial patches across 10 categories, to implement network-agnostic patch attacks (see Sec.~\ref{sec:background-attacks}). Specifically, we applied patches from the \texttt{banana} category to all images.

As a result, for our experiments, we utilized 400 clean images, 400 patch-attacked images, and a total of $42000=(8 \times 7)\times 50 \times 3 \times 5$ ($\text{class pairs}\times\text{images/attack}\times\text{attack types}\times\text{models}$) gradient-attacked images (Fig.~\ref{fig:adversarial-sample-examples}).

%

\subsection{Layer Selection}
\label{sec:setup-layers}

For our experiments, we defined two groups of layers: layers for latent space comparison and layers for concept discovery. The selected layers for all tested networks are listed in Table~\ref{tab:layers}.

\input{graphics/table_layers}

Layers for comparing latent space representations (see Sec.~\ref{sec:experiments-activations}) are evenly distributed along the depth of the model. In these layers, we assess the impact of adversarial attacks on the internal representations of the models.

For the concept discovery in adversarial attacks (see. Sec~\ref{sec:experiments-mining}, Sec~\ref{sec:experiments-adversarial-perturbation}) we use another set of layers, which are located deeper in the networks. This allows us to extract and compare high-level abstract concepts. 
This selection is based on the subjective qualitative assessment of the extracted layers.


%% file: graphics/table_layers.tex
\begin{table}[!t]

\setlength{\tabcolsep}{6pt}
\centering
\caption{Selected CNN layers for latent space comparison and 
concept discovery (f=features, l=layer).}

\label{tab:layers}

\begin{tabular}{c||c|c|c|c||c}
\hline

\multirow{2}{*}{Classifier} & \multicolumn{5}{c}{Selected layers} \\
\cline{2-6}
& \multicolumn{4}{c||}{Latent space comparison} & Concept discovery \\
\hline
\hline
VGG & f.4 & f.9 & f.14 & f.19 & f.19 \\
\hline
SqueezeNet & f.3 & f.9 & f.9 & f.12  & f.12 \\
\hline
MobileNet & f.3 & f.7 & f.16 & f.11 & f.16 \\
\hline
ResNet18 & l1.1 & l2.1 & l3.1 & l4.1 & l4.0 \\
\hline
ResNet50 & l1.1 & l2.2 & l3.4 & l4.2 & l4.2 \\
\hline

\end{tabular}
\end{table}

%% file: sections/5_experiments.tex
\section{Experimental Results}
\label{sec:experiments}

In this section, we examine the influence of AAs on sample representations within the latent space (Sec.\,\ref{sec:experiments-activations}). Subsequently, leveraging this analysis, we employ concept discovery (mining) to quantitatively and qualitatively assess the alteration of concepts before and after the attacks (Sec.\,\ref{sec:experiments-mining}). Finally, we analyze the components of adversarial perturbation through concept discovery (Sec.\,\ref{sec:experiments-adversarial-perturbation}).

\input{graphics/figure_similarity_plots_a}
\input{graphics/figure_similarity_plots_b}

\subsection{Adversarial Attacks Impact on Latent Space Representations}
\label{sec:experiments-activations}

\ResearchQuestion{What is the impact of adversarial attacks on sample embeddings within the feature space?}

To assess this, we evaluate the cosine similarities between attacked and non-attacked samples across each origin-target class pair. Figure~\ref{fig:similarity-plots-a} and Figure~\ref{fig:similarity-plots-b} show the mean curves and standard deviation intervals for the $\texttt{garbage truck} \rightarrow \texttt{banana}$, $\texttt{orange} \rightarrow \texttt{taxi}$, and $\texttt{pickup} \rightarrow \texttt{zebra}$ attacks across selected layers (Tab.~\ref{tab:layers}) of all tested networks.

Across all attack types, we observe a \enquote{snowball effect} of cosine similarity decline: the similarity diminishes exponentially as we move closer to the deeper layers, indicating a higher impact of attacks on internal representations within these layers. This effect is particularly pronounced in deeper networks containing more non-linear layers. In the final layers of ResNet50 and MobileNet, cosine similarity nearly reaches a value of $0.2$ for BIM and PGD attacks.
%
In comparison to other attacks, the C\&W attack typically induces smaller perturbations, aligning with its original intention of seeking the minimally sufficient perturbation.
%
The perturbation observed in the initial layers of the adversarial patch attack (PATCH) ($\texttt{garbage truck} \rightarrow \texttt{banana}$) resembles that of BIM and PGD attacks, yet the decline in similarity is less steep, ending in the last layer at approximately the same level as C\&W.
%
%
Based on these findings, we proceed with the following concept discovery experiment in the deep layers of the network, where activation maps are most substantially perturbed.

\Summary{Adversarial attacks cause attacked latent space representations with increasing depth to increasingly deviate from the original representations with respect to cosine similarity. This error amplification over depth is superlinear and holds across all networks, attacks, and origin-target class pairs.}

\input{graphics/figure_conept_mining}

\subsection{Concept Discovery in Adversarial Samples}
\label{sec:experiments-mining}

\ResearchQuestion{What is the impact of adversarial attacks on the main components (concepts) present in latent representations?}

In Figure~\ref{fig:concept-mining-555-468}, we showcase qualitative and quantitative outcomes of concept discovery using ICE~\cite{zhang2021invertible} and saliency-based concept comparison~\cite{mikriukov2023quantified} in $layer4.0$ of ResNet18. We focus on original \texttt{fire truck} samples and samples attacked by BIM and C\&W targeting \texttt{taxi} (PGD results omitted due to their similarity to BIM; patch attack results omitted as they lead expectedly to the discovery of the \texttt{banana} patches as a distinct concept).

\paragraph{Qualitatively,}
(1) AAs modify the \emph{concept information}, resulting in concept saliency map changes; e.g., \texttt{windshield} concept (original $\texttt{c4}$, BIM $\texttt{c0}$, and C\&W $\texttt{c4}$) highlights different areas in \nth{1} prototypes.
(2) AAs may introduce \emph{new concepts}; e.g. BIM $\texttt{c4}$ is a new spurious concept that cannot be interpreted.
(3) Additionally, a \emph{change in most similar concept prototypes} can be observed: e.g., \nth{2} prototypes of \texttt{windshield} (original $\texttt{c4}$, BIM $\texttt{c0}$, and C\&W $\texttt{c4}$) are different.

\paragraph{Quantitatively,}
we observe
(1) changes of values in concept similarity matrices (Fig.~\ref{fig:concept-mining-555-468}, bottom);
(2) alterations of concept importance (e.g., concept \texttt{cabin bottom} weights; e.g., original $\texttt{c0: -7.61}$, BIM $\texttt{c3: -24.97}$, and C\&W $\texttt{c0: -27.45}$), which may result in
(3) concept rank shuffling (e.g., concept \texttt{cabin bottom}: original  $\texttt{c0: \nth{1}}$, BIM $\texttt{c3: \nth{3}}$, and C\&W $\texttt{c0: \nth{3}}$).

To find similar concepts (counterparts), columns of the concept similarity matrix are permuted to maximize the sum of the matrix diagonal. The diagonal values indicate the magnitude of similarity between these counterparts. Stronger attacks result in more pronounced perturbations, leading to the emergence of new concepts or the disappearance of old ones. Some concepts may have very weak counterparts, which can be interpreted as no counterpart or that \enquote{concept was changed}: e.g., BIM $\texttt{c4}$ (Fig.~\ref{fig:concept-mining-555-468}, bottom left matrix).

\input{graphics/figure_mining_concept_changes}

To measure it, for all discovered concepts for every origin-target attack pair, we threshold to matrix diagonal values and estimate mean and 99\% confidence intervals of case counts when the value is lower than used threshold. These results for all networks and threshold values of 75, 50, and 25 are depicted in Figure~\ref{fig:mining-concept-changes}. These results represent the average number of \enquote{concept change} occurrences under the adversarial attacks, serving as a measure of attack strength. Numerically, we observe similar behavior between BIM and PGD. C\&W, which induces the smallest perturbations among the tested attacks, results in the lowest number of concept changes across all networks.


In some rare instances, depending on the chosen threshold, no concepts may be modified. However, the impact of the AA can still be observed through the evaluation of changes in concept weights, concept ranks, or concept prototypes. For instance, in Fig.~\ref{fig:concept-mining-555-468}, each C\&W attack concept (bottom right matrix) corresponds to a counterpart for which the concept weight is known. To estimate changes in weights and concept ranks, Spearman's rank correlation and Pearson's correlation can be measured. For the C\&W attack and the setup illustrated in Figure~\ref{fig:mining-concept-changes}, these correlations are $0.6$ and $0.76$, respectively, enabling the detection of concept changes.


\Summary{Comparing concept composition in clean sample representations to ones in adversarially attacked counterparts shows: AAs modify all of concept saliency maps, concept weights respectively ranking, and concept similarities; and even replace concepts in case of strong attacks.}

\subsection{Concept Analysis of Adversarial Perturbation}
\label{sec:experiments-adversarial-perturbation}

Denote by $f = f_{L \to}\circ f_{\to L}$ be the decomposition of the CNN into the part up to layer $L$ an the part from $L$ onwards. The representation of an adversarial perturbation $\delta$ for a sample $x$ in $L$'s latent space can be defined as
\begin{align}
    \label{eq:experiments-adversarial-perturbation}
    \tilde{\delta} = \delta_{x,L} \coloneqq f_{\to L}(x + \delta) -  f_{\to L}(x)
\end{align}

\subsubsection{Information Distribution in Adversarial Perturbation.}

Here, our goal is to discover whether the information within adversarial perturbation $\tilde{\delta}$ has dominant directions or/and has an imbalanced distribution. In other words:
\ResearchQuestion{Can an adversarial perturbation in latent space be efficiently represented using linear components? What is the minimum number of components needed?}

For this, we quantify the information distribution across channel dimensions (see Sec.~\ref{sec:background-concept-discovery}) of $\tilde{\delta}$ with PCA decomposition and assess the cumulative variance explained by obtained components. We determine the minimum part of PCA components (relative to the total number of components in a layer) needed to preserve a defined fraction of the variance cumulatively. The total number is equal to the number of convolutional filters in each discovery layer (see Table~\ref{tab:layers}). 

In Table~\ref{tab:pca-variance}, we present the mean and standard deviations of these ratios, expressed as percentages (\%). Across selected layers (Tab.~\ref{tab:layers} \enquote{Concept Discovery}) of all networks and attack types, we observe an uneven distribution of information among components: depending on the model and attack type, (1) 0.3\% to 11.5\% of components are sufficient to retain 50\% of the whole variance; and (2) preserving 7.6\% to 58.3\% of components, we can explain 90\% of variance. (3) In selected layers with a larger amount of filters, like in MobileNet (960) and ResNet50 (2048), 99\% of variance is explained by 27.1\% (C\&W of ResNet50) to 77.9\% (BIM, PGD of MobileNet) of components, whereas in models with fewer filters (VGG, SqueezeNet, ResNet18: 512 filters in all selected layers) 99\% of variance is explained by 66.9\% (PATCH of VGG) to 91.5\% (BIM, PGD of ResNet18).

\input{graphics/table_pca_variance}

BIM and PGD attacks create larger perturbations and consistently require the highest portions of components to explain $\tilde{\delta}$ variance properly. In contrast, the adversarial patch (PATCH) attack necessitates the smallest portion of components. The behavior of the C\&W attack varies across models: in some cases, its results align with those of BIM and PGD, while in others, they resemble the outcomes of the PATCH attack.

\Summary{The majority of $\tilde{\delta}$ information is concentrated in a small subset of linear components (concepts), i.e., $\tilde{\delta}$ is built from few meaningful directions in the feature space.}

\subsubsection{Concept Discovery in Adversarial Perturbations.}
From the previous experiments we know that latent space perturbations originating from a given attack do give rise to a global linear decomposition.
For the discovery of concepts in $\tilde{\delta}$ we next utilize NMF, as it learns more meaningful directions than other decomposition methods~\cite{zhang2021invertible}. To ensure the non-negativity constraint in NMF, like in ICE~\cite{zhang2021invertible}, we apply $\tilde{\delta}^+ = \texttt{ReLu}(\tilde{\delta})$. Even though negative perturbation information is lost, this approach still results in successful concept discovery. 
Following the application of NMF (see Sec.~\ref{sec:background-concept-discovery}) to $\tilde{\delta}^+$, we obtain $M^+ \in \Reals^{k \times c}$.
For ease of notation let $m_i={M_i^+}/{\|M_i^+\|}$ be the normalized $i$-th concept vector.
Given the concepts, our next question is:
\ResearchQuestion{Are single linear components of adversarial perturbations in latent space sufficient to reproduce the attack? In other words: Can adversarial attacks be described by linear translations in latent space?}


Recall that the $m$-component of $\tilde{\delta}$ for a vector $m$ (i.e., the amount by which $\tilde{\delta}$ points into direction $m$) is:
\begin{align}
\SwapAboveDisplaySkip
\text{proj}_{m} \big( \tilde{\delta} \big) \coloneqq \big(\tilde{\delta} \circ m\big) m
\label{eq:experiments-projection-onto-nmf-component}
\end{align}
We would now like to compare on the outputs the effect of slowly applying $\tilde{\delta}$ with the effect of slowly applying any of its components $m_i$.
Slow application is here realized by linearly interpolating between $\tilde{x}\coloneqq f_{\to L}(x)$ and $\tilde{x} + \tilde{\delta} = f_{\to L}(x+\delta)$ via $\tilde{x} + \gamma \tilde{\delta}$, $\gamma\in[0,1]$.
 To ensure comparability of the linear interpolations, we interpolate each $m_i$ via $\tilde{x}+\gamma \text{proj}_{m_i}(\tilde{\delta})$. This ensures that at most $\tilde{\delta}$ is applied to $\tilde{x}$.
Finally, we estimate and compare the influence of the interpolations on prediction confidences of original/target class $\text{cls}$ for components $m_i, i\leq k$ at the same $\gamma\in[0,1]$:
\begin{gather}
%
\SwapAboveDisplaySkip
f_{L\to}\left(\tilde{x} + \gamma\,\tilde{\delta}\right)_{\text{cls}} 
\quad \text{versus}\quad
f_{L\to}\left(\tilde{x} + \gamma\,\text{proj}_{m_i}(\tilde{\delta})\right)_{\text{cls}} 
\label{eq:experiments-prediction-from-adversarial-perturbation}
\end{gather}
When $\gamma=0$, the prediction is made for the original image, and at $\gamma=1$, for the image attacked by full perturbation respectively its full $i$th component. 


In Figure~\ref{fig:logits}, we display dependency graphs depicting the averaged --- for all related test samples --- true class $y$ (solid lines) and attacker target class $y'$ (dashed lines) output confidence corresponding to the magnitude $\gamma$ of $\tilde{\delta}$ (\blackLine) and $k=3$ NMF component projections (\redLine, \greenLine, \blueLine). Results are shown for every techniques of attacking $\texttt{pickup} \rightarrow \texttt{banana}$ and for three models (remainder skipped due to space constraints; behavior was similar).

\input{graphics/figure_logits}

\paragraph{Walking towards original perturbation:}
In the case of original perturbation $\tilde{\delta}$ (\blackLine{} lines), we observe that (1) gradient attacks, which create large perturbations (PGD, BIM), reach target confidences $f_{L\to}(\tilde{x} + \gamma\,\tilde{\delta})_{\text{target}} \approx 1.0$ at $\gamma = 1$ and and maintain this value thereafter, while (2) the low perturbation attack (C\&W) does not reach a value close to 1.0; however, it can be further amplified by increasing $\gamma > 1$. (3) PATCH attack behaves similarly to C\&W, but in some cases the target confidences again drop at high $\gamma$ values (see VGG). In other models and origin-target combinations of AAs, the observed behavior was consistent.

\paragraph{Walking towards linear perturbation components:}
For $k=3$ concepts discovered with NMF decomposition, we observe that (1) one or several NMF components (typically the two most prominent ones \redLine{} and \greenLine) in each of the demonstrated cases lead to successful attacks, i.e., increasing the target class confidences and decreasing original class ones. However, the decrease of original and growth of target class confidences are slower than these for $\tilde{\delta}$ (\blackLine). (2) Discovered concepts behave differently; the \blueLine{} component usually contributes positively to original class $y$ probability, reinforcing it. This may be caused by partial information loss resulting from the NMF non-negativity constraint. The observed behavior of NMF concepts is consistent across all networks and adversarial attack origin-target combinations.
We observed a similar behavior for concepts discovered in PCA; however, the impact of $\gamma$ was weaker, meaning that the decrease in $y$ and the growth in $y'$ were slower. Due to this similarity, we do not present visual results.

\Summary{A given adversarial attack can be replaced by walking in latent space into the direction of the most prominent linear component(s) of its adversarial perturbations.}

\input{graphics/figure_nmf_similarity_one_orig_a}
\input{graphics/figure_nmf_similarity_one_orig_b}

\input{graphics/figure_nmf_similarity_resnet50_468}
  
\subsubsection{Similarity of Concepts Discovered in Adversarial Perturbation.}
Considering the observed similarity in the behavior of perturbation components (\redLine, \greenLine, \blueLine) and of the original perturbation $\tilde{\delta}$ (\blackLine) across all test cases, it is a valid assumption that we detected similar concepts in tested AAs, possibly with varying magnitude depending on the attack technique. 
In other words:
\ResearchQuestion{Do adversarial attacks comprise concept vectors of similar directions in the feature space?}

To investigate the similitude of concept vectors discovered in $\tilde{\delta}$ across different attack techniques we estimate cosine similarities (see Eq.~\ref{eq:cosine-similarity}) for each pair of vectors. In Figs.\,\ref{fig:nmf-sim-one-orig-a} and \ref{fig:nmf-sim-one-orig-b} we present the similarity clustermaps\footnote[4]{\url{https://seaborn.pydata.org/generated/seaborn.clustermap}} (clustered heatmaps) of concept vectors $M^+$ discovered with NMF for all tested attack techniques and different origin-target attack pairs.

\paragraph{Similarities across attack techniques:}
We observe that (1) concepts discovered in different attack techniques for the same are similar (co-directed) and form clusters (visible as groups of brighter pixels). For instance, PA1, C1, B1, and P1 group of ResNet18 (for attacking $\texttt{zebra} \rightarrow \texttt{banana}$) contains one vector of each category. 
(2) At least one of such groups is observed per clustermap. Similar behavior was observed in other origin-target attack pairs not visualized here.
(3) Adversarial patch attack vectors are the most distinct, as evident in the $\texttt{zebra} \rightarrow \texttt{banana}$ column. This distinction can be attributed to the different nature of the attack compared to the other gradient-based attack techniques.

\paragraph{Similarities across origin classes:}
We further expand the pairwise concept vector comparison to vectors originating from any category, only fixing the attack target class. In other words, we investigate whether the learned directions of concept vectors $M^+$ discovered with NMF are purely \emph{target-specific}, respectively in how far they are agnostic to attack type and original true class. In Fig.\,\ref{fig:nmf-sim-rn50-468} we showcase a concept similarity clustermap for ResNet50, where concept vectors are targeting the \texttt{taxi} class ($\texttt{any} \rightarrow \texttt{taxi}$). Similar to notations in Fig.\,\ref{fig:nmf-sim-one-orig-a}, rows and columns are indexed as \texttt{OriginClassId-AttackType-ConceptId}. Here, the additional \texttt{OriginClassId} represents the ImageNet class ID of the AA origin.
From such results of \emph{target-specific} comparison of concept vectors, we observe two large groups of similar (co-directed) concepts (approximately $20 \times 20$ pixels each) and several smaller groups (around $3 \times 3$ pixels). The large groups include concept vectors of attacks originating from all tested supercategories (\texttt{vehicle}, \texttt{animal}, and \texttt{fruit}) and targeting \texttt{taxi} class. Similar results were observed for different attack targets and models: at least one large cluster of concept vectors was observed.

\Summary{Our results imply that adversarial attacks comprise concept vectors of \emph{target-specific} direction (albeit with varying magnitudes depending on the AA perturbation strength). In particular, AAs can be characterized by directions in feature space that are mostly agnostic to attack technique and attack origin class.}
This knowledge can be further leveraged in the explainable design of adversarial attacks and defenses against them.

%% file: graphics/figure_similarity_plots_a.tex
\begin{figure}[!b]
  \centering
  \includegraphics[width=1.0\linewidth]{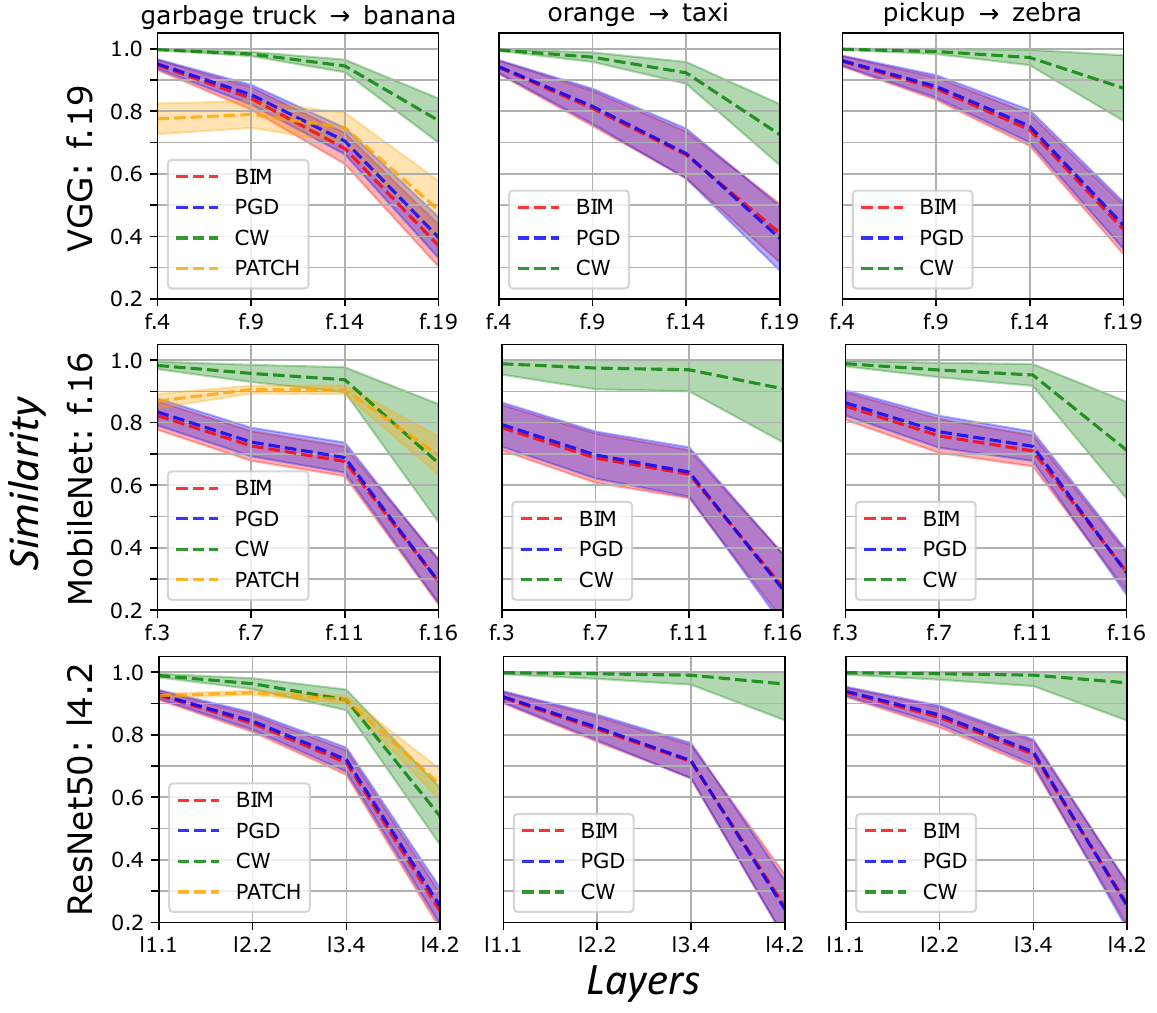}
  \caption{Mean and standard deviation values of cosine similarities for original and attacked activation maps of test samples
  for several attacks.}
  \label{fig:similarity-plots-a}
\end{figure}

%% file: graphics/figure_similarity_plots_b.tex
\begin{figure}[!t]
  \centering
  \includegraphics[width=1.0\linewidth]{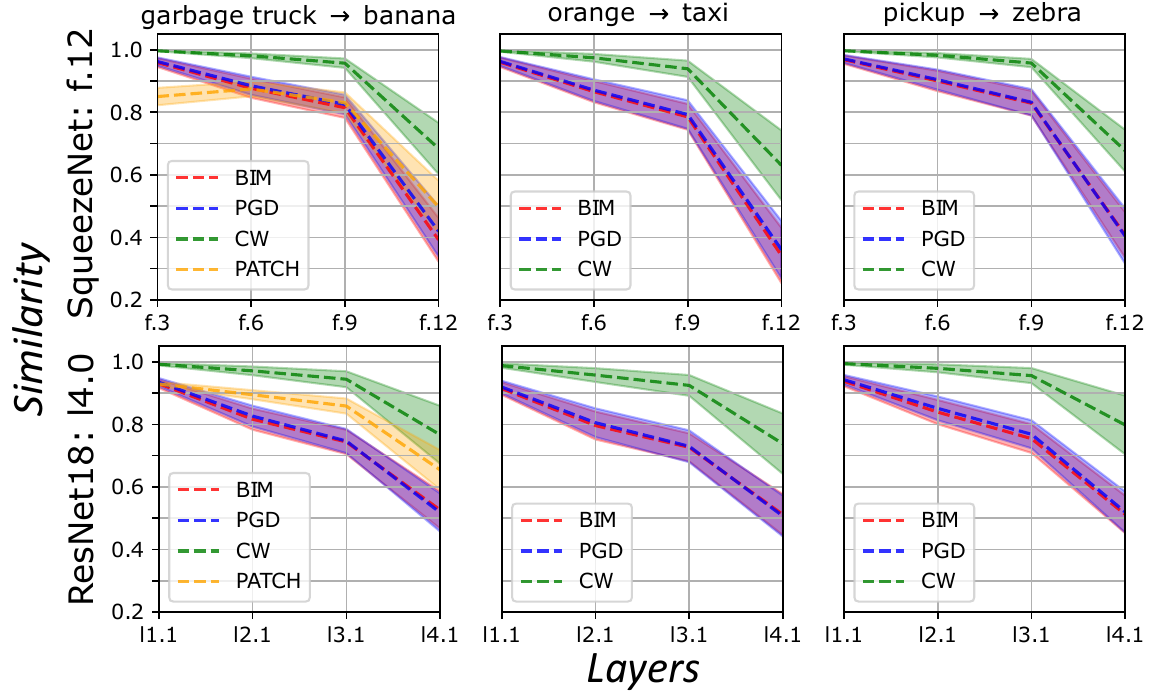}
  \caption{Mean and standard deviation values of cosine similarities for original and attacked activation maps of test samples
  for several attacks.}
  \label{fig:similarity-plots-b}
\end{figure}

%% file: graphics/figure_conept_mining.tex
\begin{figure}[!b]
  \centering
  \includegraphics[width=1.0\linewidth]{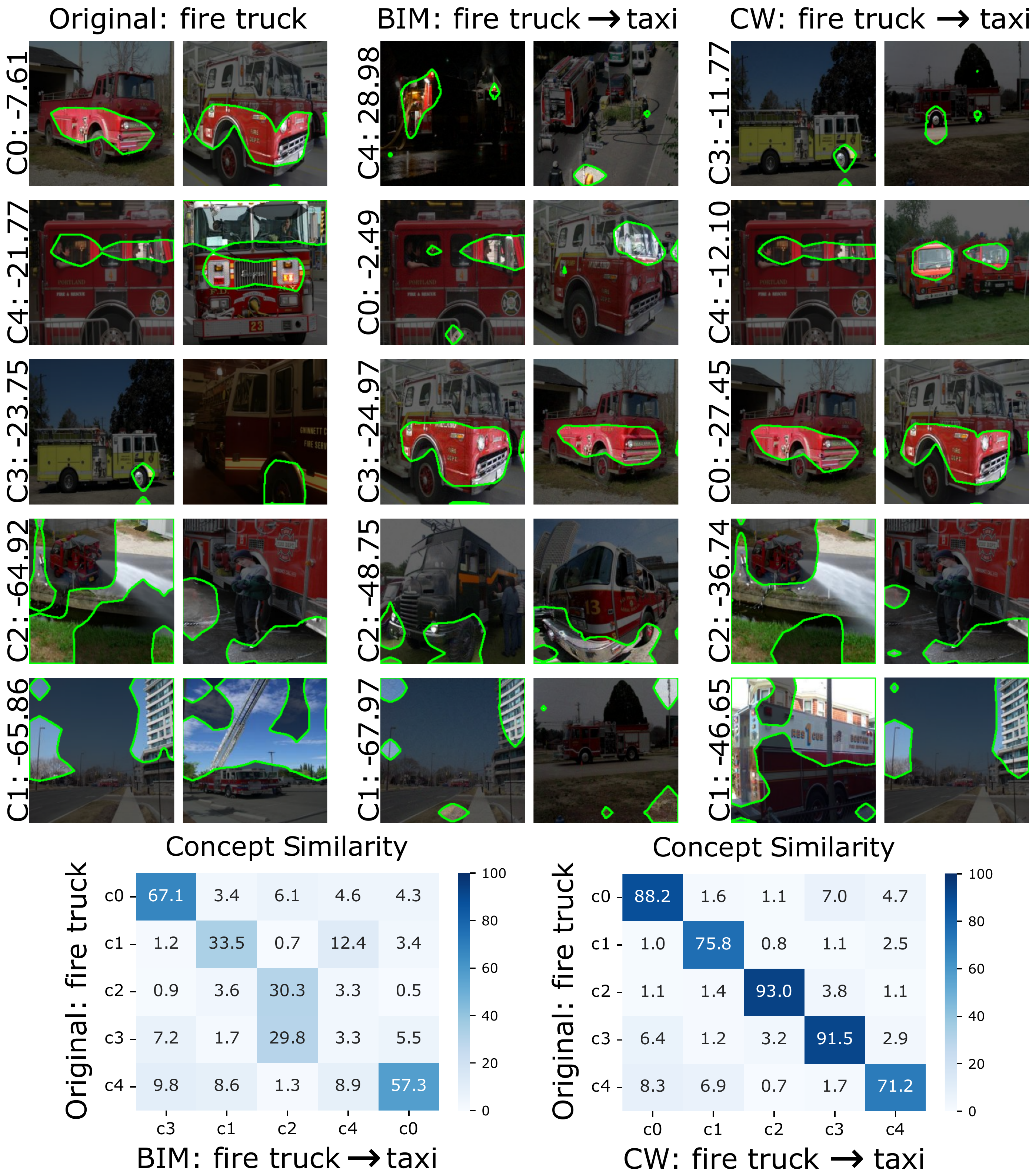}
  \caption{Concept mining results for BIM and C\&W ($\texttt{fire truck} \rightarrow \texttt{taxi}$) attacks in layer $layer4.0$ of ResNet18.
  \emph{Top:} pairs of top-2 most relevant prototypes of discovered concepts {\sffamily cX} with rank {\sffamily X} and concept weights (importances);
  \emph{Bottom:} discovered concept similarities (Sec.~\ref{sec:background-comparison}) for original vs. BIM \emph{(left)} and original vs. C\&W \emph{(right)} }
  \label{fig:concept-mining-555-468}
\end{figure}

%% file: graphics/figure_mining_concept_changes.tex
\begin{figure}[!b]
  \centering
  \includegraphics[width=1.0\linewidth]{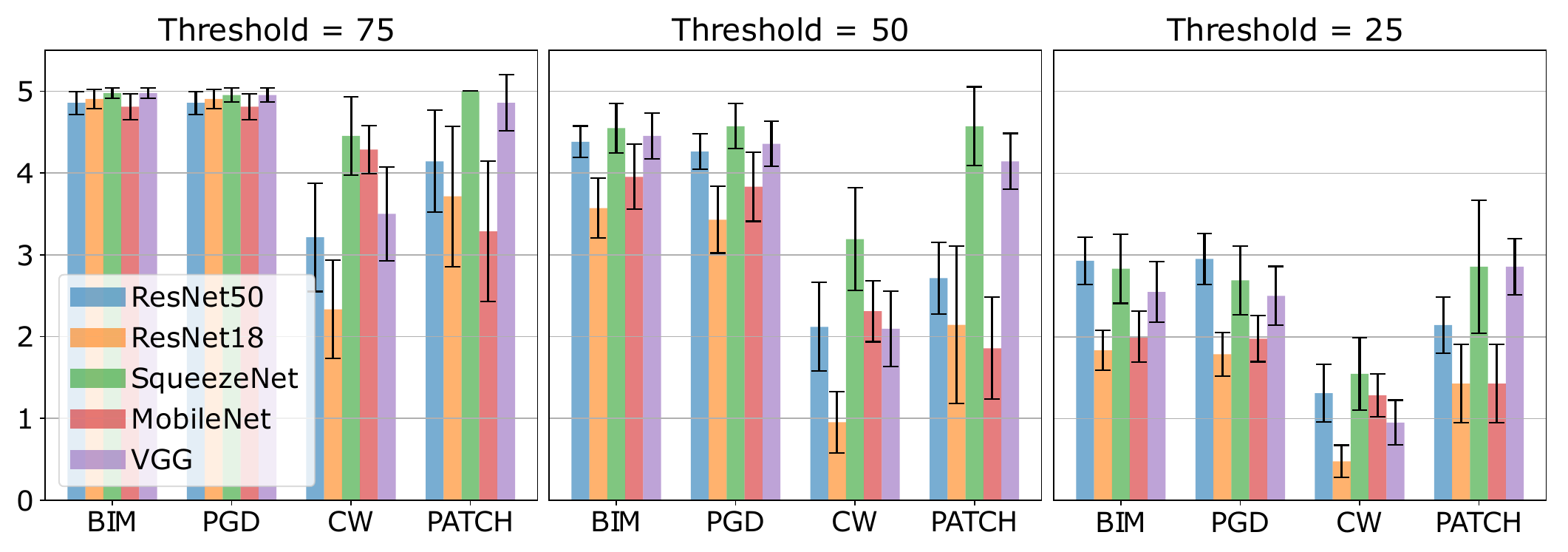}
  \caption{Mean numbers of \enquote{concept changes} with 99\% confidence intervals for threshold values 75, 50, and 25 in all tested models for tested adversarial attacks.}
  \label{fig:mining-concept-changes}
\end{figure}

%% file: graphics/table_pca_variance.tex
\begin{table}[!t]

\setlength{\tabcolsep}{2pt}
\centering
\caption{Mean and standard deviation ratios (in \%) of preserved PCA components relative to the total number of components in layer (\#\texttt{filters}) required to retain a specified amount of variance.}

\label{tab:pca-variance}

\fontsize{8pt}{10pt}\selectfont
\begin{tabular}{c|c||c|c|c|c|c|c}

\hline
\multirow{2}{*}{\makecell{Model: Layer \\ (\#\texttt{filters})}} & \multirow{2}{*}{Attack} & \multicolumn{6}{c}{Retained Variance} \\
\cline{3-8}
                                                 & & 50\% & 70\% & 80\% & 90\% & 95 \% & 99 \% \\
\hline
\hline
\multirow{4}{*}{\makecell{VGG: f.19\\(512)}}
                     & BIM & 8.5±1.1 & 22.1±1.9 & 34.1±2.2 & 53.1±2.5 & 68.2±2.5 & 88.4±1.8 \\
                     & PGD & 8.6±1.2 & 22.3±2.0 & 34.3±2.3 & 53.3±2.5 & 68.4±2.4 & 88.5±1.7 \\
                     & CW & 6.5±1.5 & 18.0±2.8 & 28.9±3.6 & 47.4±4.2 & 63.2±4.0 & 86.1±2.4 \\
                     & PATCH & 0.4±0.1 & 2.2±0.4 & 6.6±1.0 & 19.6±1.6 & 35.3±1.8 & 66.9±1.7 \\

\hline
\multirow{4}{*}{\makecell{SqueezeNet: f.12\\(512)}}

                            & BIM & 8.9±1.2 & 22.4±1.8 & 34.2±2.0 & 53.2±2.0 & 68.6±1.7 & 89.4±0.9 \\
                            & PGD & 8.9±1.1 & 22.4±1.7 & 34.3±2.0 & 53.3±1.9 & 68.6±1.6 & 89.4±0.9 \\
                            & CW & 9.4±1.8 & 23.2±2.6 & 35.0±2.8 & 53.9±2.7 & 69.1±2.2 & 89.7±1.1 \\
                            & PATCH & 0.9±0.2 & 4.9±0.6 & 11.1±1.1 & 25.8±1.5 & 41.7±1.9 & 72.5±1.4 \\
                            
\hline                  
\multirow{4}{*}{\makecell{MobileNet: f.16\\(960)}}

                           & BIM & 3.7±0.6 & 11.5±1.3 & 19.8±1.7 & 35.6±2.0 & 50.9±2.0 & 77.9±1.4 \\
                           & PGD & 3.8±0.6 & 11.6±1.2 & 19.9±1.6 & 35.7±1.9 & 51.0±1.9 & 77.9±1.3 \\
                           & CW & 2.3±0.7 & 7.7±2.0 & 13.9±3.4 & 26.5±6.0 & 39.5±8.6 & 65.4±13.4 \\
                           & PATCH & 0.7±0.1 & 3.6±0.5 & 7.7±1.0 & 17.9±1.8 & 30.2±2.2 & 60.0±2.4 \\
                           
\hline
\multirow{4}{*}{\makecell{ResNet18: l4.0\\(512)}}

                          & BIM & 11.5±0.7 & 26.9±1.0 & 39.4±1.1 & 58.3±1.1 & 72.8±1.0 & 91.5±0.6 \\
                          & PGD & 11.5±0.8 & 26.9±1.1 & 39.4±1.2 & 58.3±1.2 & 72.9±1.0 & 91.5±0.6 \\
                          & CW & 10.4±2.0 & 24.0±3.9 & 35.2±5.3 & 52.8±6.9 & 67.2±7.6 & 87.7±7.0 \\
                          & PATCH & 1.1±0.2 & 7.5±0.8 & 15.7±1.0 & 32.1±1.2 & 48.3±1.2 & 77.3±1.0 \\

\hline                           
\multirow{4}{*}{\makecell{ResNet50: l4.2\\(2048)}}

                          & BIM & 2.6±0.4 & 8.0±0.7 & 13.4±0.9 & 23.8±1.2 & 34.3±1.4 & 56.2±1.8 \\
                          & PGD & 2.7±0.4 & 8.0±0.7 & 13.4±0.9 & 23.7±1.2 & 34.2±1.4 & 56.0±1.6 \\
                          & CW & 1.3±0.7 & 3.8±2.1 & 6.6±3.5 & 12.1±6.3 & 17.9±9.5 & 30.5±16.7 \\
                          & PATCH & 0.3±0.1 & 1.4±0.3 & 3.2±0.5 & 7.6±0.9 & 13.0±1.3 & 27.1±1.8 \\
\hline  

\end{tabular}
\end{table}

%% file: graphics/figure_logits.tex
\begin{figure}[!t]
  \centering
  \includegraphics[width=1.0\linewidth]{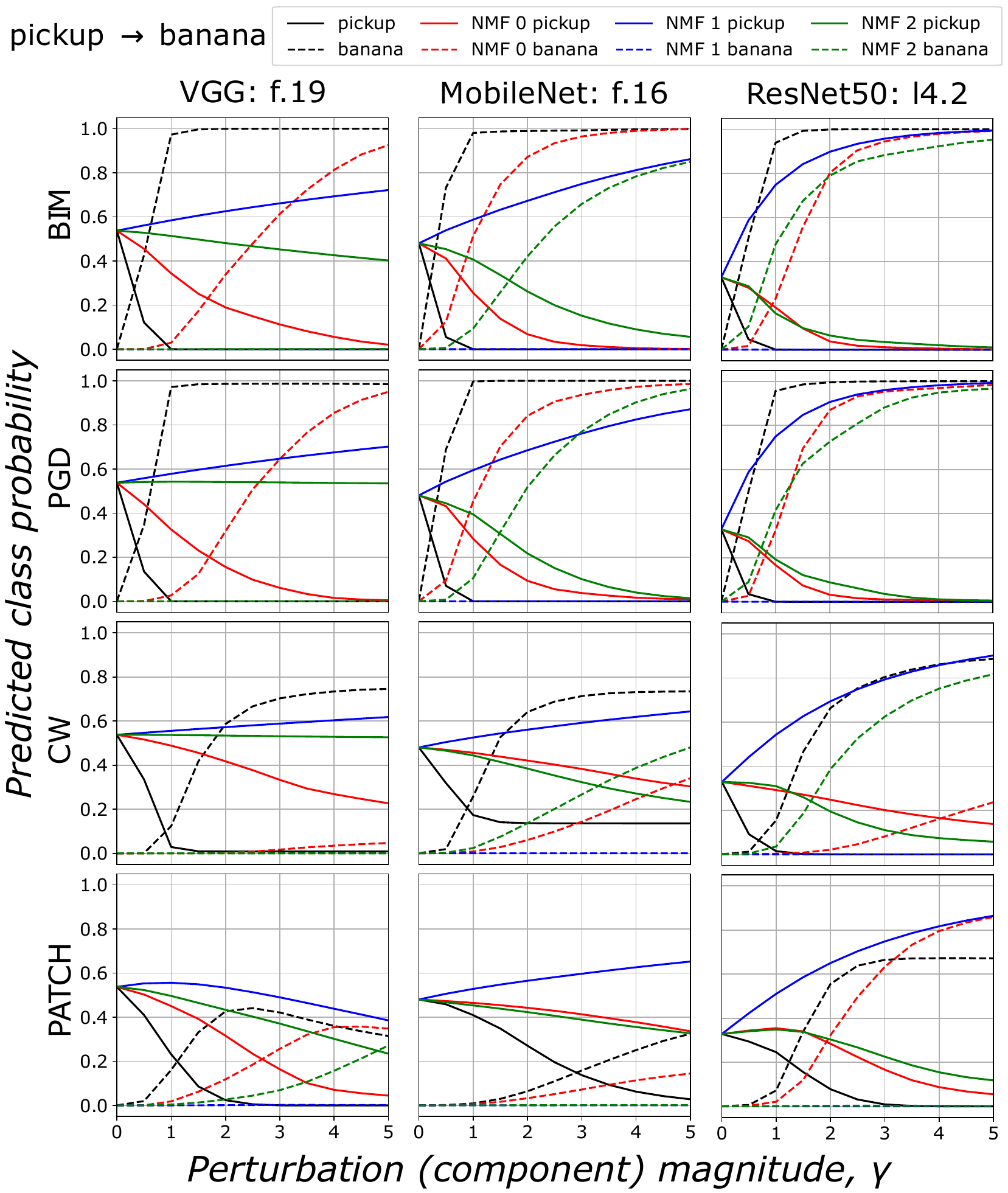}
  \caption{The dependency of original and target class probabilities on the adversarial perturbation magnitude $\gamma$ averaged for all test samples. Original adversarial perturbation (\blackLine) is compared to perturbation projected onto 3 concepts discovered with NMF (\redLine, \greenLine, \blueLine).}
  \label{fig:logits}
\end{figure}

%% file: graphics/figure_nmf_similarity_one_orig_a.tex
\begin{figure}[!b]
  \centering
  \includegraphics[width=1.0\linewidth]{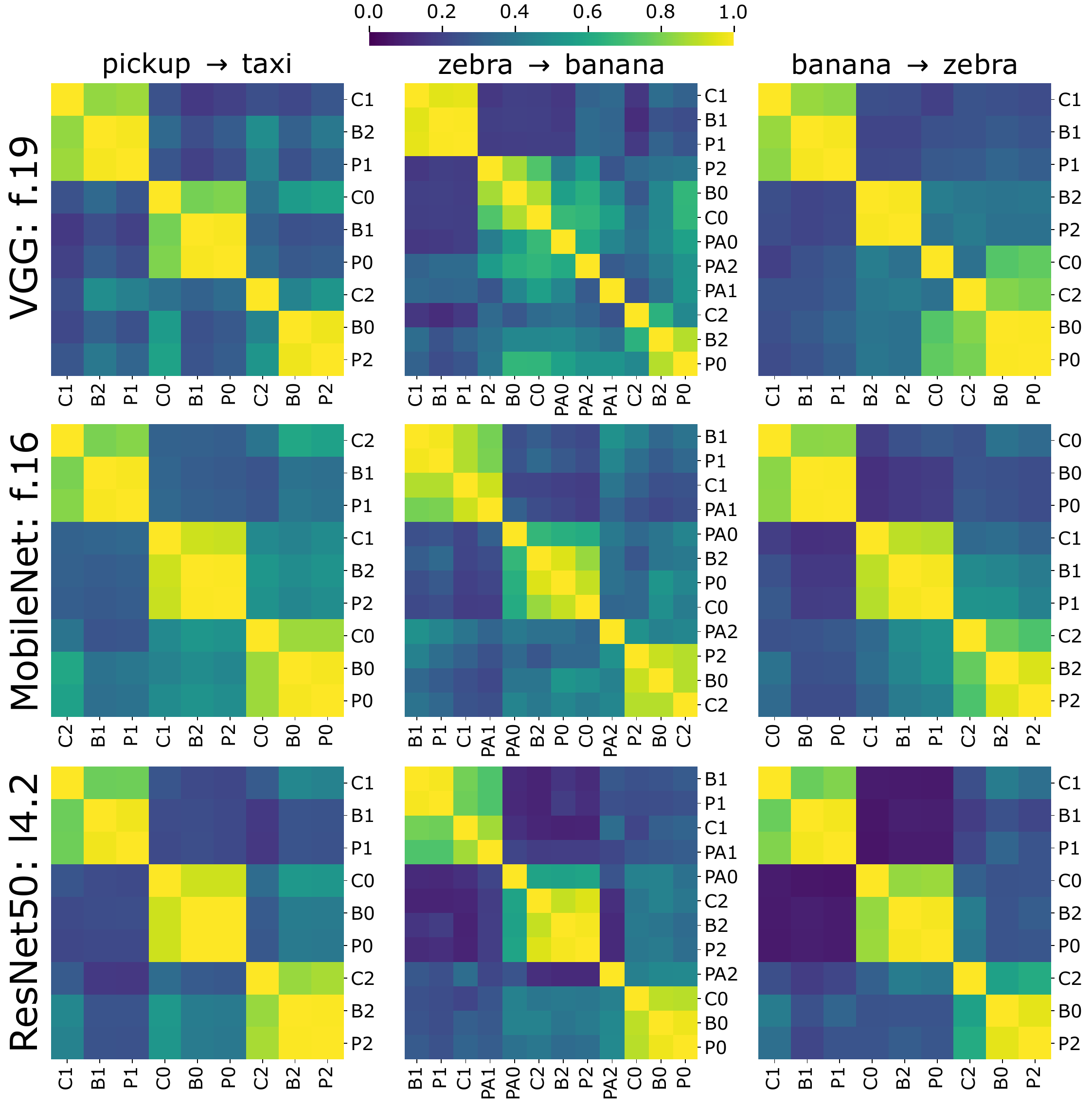}
  \caption{Similarity heatmaps of concepts (3 per attack) discovered in adversarial perturbations with NMF. Concepts are denoted as \texttt{AttackType-ConceptId} pairs (Attack types: $\text{B=BIM, P=PGD, C=C\&W, PA=Patch Attack}$).}
  \label{fig:nmf-sim-one-orig-a}
\end{figure}

%% file: graphics/figure_nmf_similarity_one_orig_b.tex
\begin{figure}[!t]
  \centering
  \includegraphics[width=1.0\linewidth]{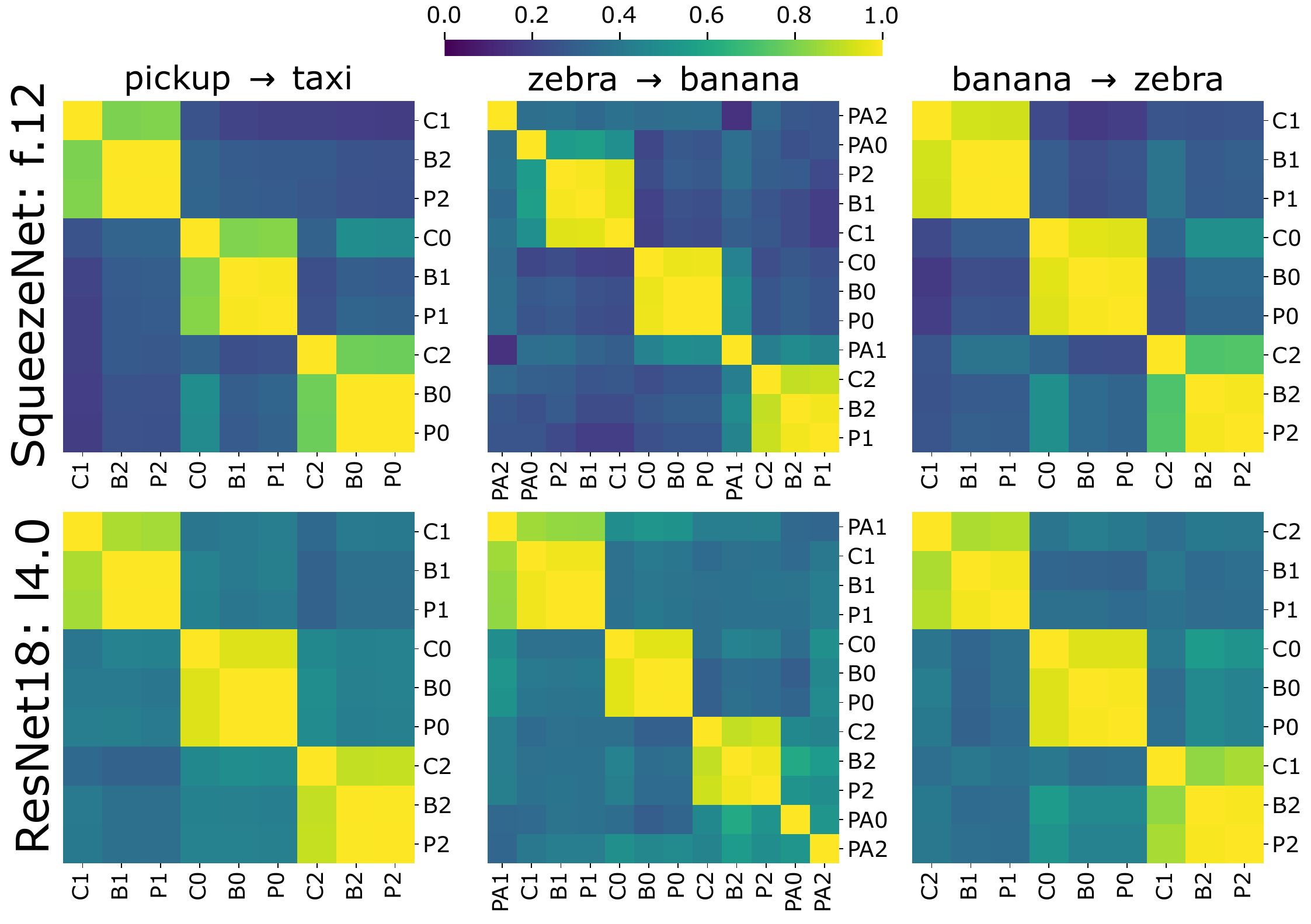}
  \caption{Similarity heatmaps of concepts (3 per attack) discovered in adversarial perturbations with NMF. Concepts are denoted as \texttt{AttackType-ConceptId} pairs (Attack types: $\text{B=BIM, P=PGD, C=C\&W, PA=Patch Attack}$).}
  \label{fig:nmf-sim-one-orig-b}
\end{figure}

%% file: graphics/figure_nmf_similarity_resnet50_468.tex
\begin{figure}[!b]
  \centering
  \includegraphics[width=1.0\linewidth]{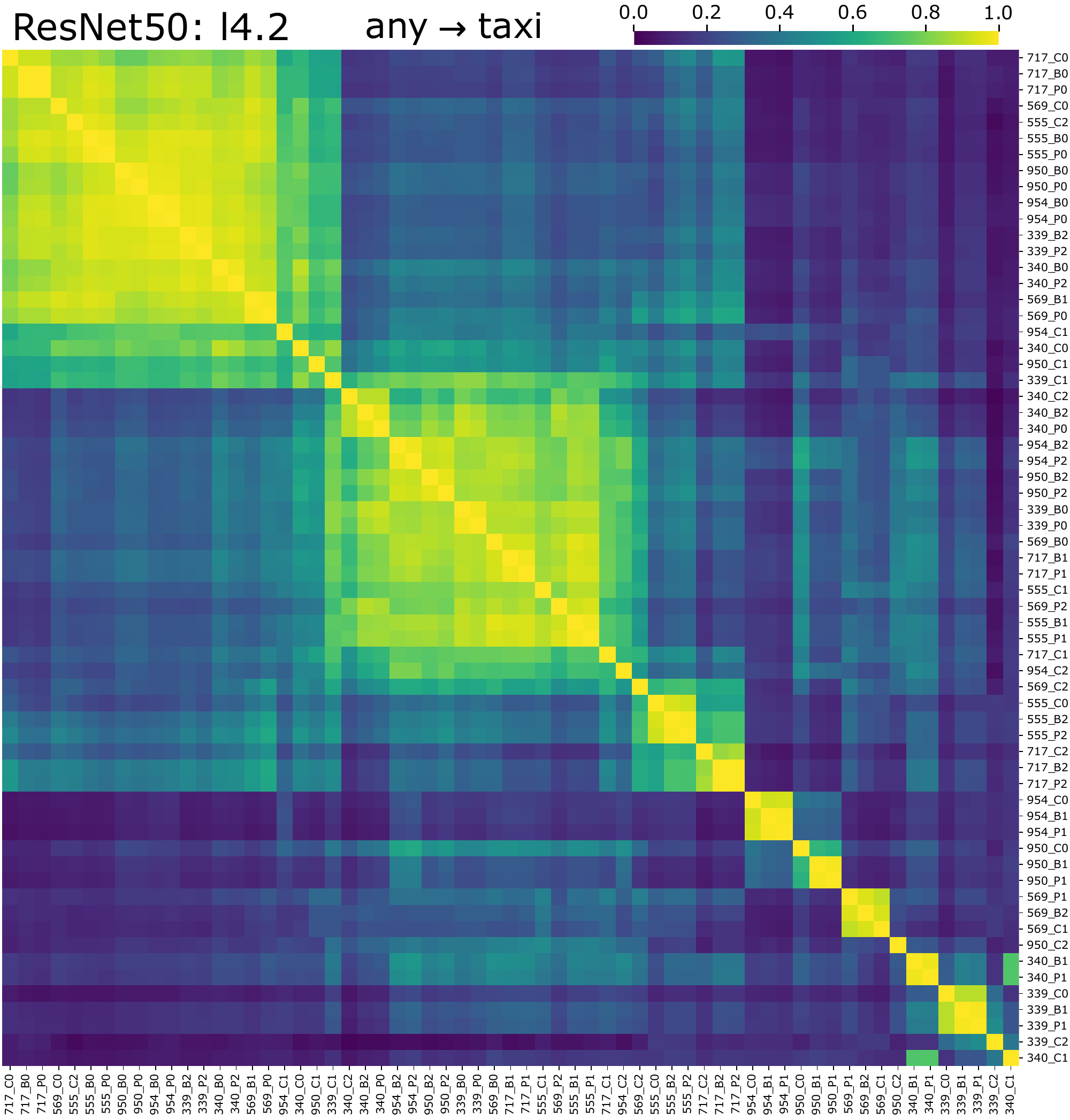}
  \caption{Similarity of all concepts discovered with NMF in adversarial perturbations aiming \enquote{taxi} class in $layer4.2$ of ResNet50. Concepts are denoted as \texttt{OriginClassId-AttackType-ConceptId} pairs (Attack types: $\text{B=BIM, P=PGD, C=C\&W}$).}
  \label{fig:nmf-sim-rn50-468}
\end{figure}

%% file: sections/6_conclusion.tex
\section{Conclusion and Outlook}
\label{sec:conclusion}




This work for the first time conducted an in-depth analysis of the influence of AAs on the concepts learned by CNNs using concept-based XAI techniques. Through experiments our results revealed that AAs induce substantial alterations in the concept composition within the feature space, introducing new concepts or modifying existing ones. Remarkably, we demonstrated that the adversarial perturbation itself can be decomposed into a set of concepts, a subset of which is sufficient to reproduce the attack's success. Furthermore, we discovered that different AAs learn similar concept vectors, and that these vectors are only specific to the attack target class irrespective of the origin class.

These findings provide valuable insights into the nature of AAs and their impact on learned representations, paving the way for the development of more robust and interpretable deep learning models, as well as effective defenses against adversarial threats.

\subsubsection{Limitations:} Our experiments focused on pairwise training scenarios and targeted white-box attacks. Extending the analysis to non-targeted attacks, further black-box attacks, and universal attacks as well as other model types is needed in future to broaden our understanding of AAs' impact on learned concepts.

\subsubsection{Future Work on Countering Attacks:}
It will be interesting to see whether the concept-level patterns and alterations induced by AAs are useful for (real-time) detection of adversaries. Additionally, our findings on how to globally represent AAs as linear shifts in latent space could inform the design of adversarial elimination techniques that aim to remove or mitigate the impact of adversarial concepts within the learned representations.

Another intriguing direction is the explainable design of AAs themselves. By leveraging our understanding of the target-specific nature of adversarial concepts, novel attack strategies that optimize for specific concept directions could be developed, potentially leading to more effective and robust adversarial examples---which again can serve in adversarial training for creating super-robust models.

A potential next iteration of our approach is the exploration of non-linear AAs that manipulate the interactions between concepts rather than solely targeting individual concept directions. Such attacks could potentially circumvent existing defenses and uncover important challenges to the robustness of deep learning models.

In conclusion, our study has demonstrated the value of leveraging explainable AI techniques to gain insights into the impact of AAs on learned representations and concepts within deep neural networks. By bridging the gap between adversarial robustness and CNN latent space interpretability, we hope to have paved the way towards more reliable and trustworthy AI systems capable of withstanding adversarial threats whilst providing transparent and interpretable decision-making processes.
